\documentclass[lettersize,journal]{IEEEtran}
\usepackage{amsmath,amsfonts,amssymb}
\usepackage{algorithmic}
\usepackage{algorithm}
\usepackage{array}
\usepackage[caption=false,font=normalsize,labelfont=sf,textfont=sf]{subfig}
\usepackage{textcomp}
\usepackage{stfloats}
\usepackage{url}
\usepackage{verbatim}
\usepackage{graphicx}
\usepackage[numbers]{natbib}
\usepackage{booktabs}
\usepackage{multirow}
\usepackage{makecell}
\usepackage{bbding}
\usepackage[switch,columnwise]{lineno}
\usepackage[colorlinks]{hyperref}

\begin{document}
% \linenumbers

\title{InfRS: Incremental Few-shot Object Detection in Remote Sensing Images}

\author{Wuzhou~Li,
        Jiawei~Zhou,
        Xiang~Li,
        Yi~Cao,
        Guang~Jin,
        and~Xuemin~Zhang
        % <-this % stops a space‘’

\thanks{This work was supported in part by the Fundamental Research Funds for the Central Universities (Grant NO.2042022DX0001) and in part by the Hubei Luojia Laboratory Special Fund (Grant NO.220100036). \textit{(Corresponding author: Xuemin~Zhang.)}}% <-this % stops a space
\thanks{Wuzhou~Li, Guang~Jin, and Xuemin~Zhang are with the School of Remote Sensing and Information Engineering, Wuhan University, Wuhan 430072, China (email: liwuzhou1108@whu.edu.cn; jinguang5812@sina.com; zhangxuemin@whu.edu.cn).}%
\thanks{Jiawei~Zhou and Yi~Cao are with the Electronic Information School, Wuhan University, Wuhan 430072, China (email: zhoujw@whu.edu.cn; caoyi@whu.edu.cn).}%
\thanks{Xiang~Li is with the Electrical and Computer Engineering, King Abdullah University of Science and Technology, Thuwal 23955, Saudi Arabia (email: xiangli92@ieee.org).}
}

%The paper headers
\markboth{Journal of \LaTeX\ Class Files,~Vol.~14, No.~8, August~2021}%
{Li \MakeLowercase{\textit{et al.}}: InfRS: Incremental Few-shot Object Detection in Remote Sensing Images}

\IEEEpubid{0000--0000/00\$00.00~\copyright~2021 IEEE}
% % Remember, if you use this you must call \IEEEpubidadjcol in the second
% % column for its text to clear the IEEEpubid mark.

\maketitle

\begin{abstract}
Few-shot detection in remote sensing images has witnessed significant advancements recently. Despite these progresses, the capacity for continuous conceptual learning still poses a significant challenge to existing methodologies. In this paper, we explore the intricate task of incremental few-shot object detection (iFSOD) in remote sensing images. We present a pioneering transfer-learning-based technique, termed InfRS, designed to enable the incremental learning of novel classes using a restricted set of examples, while simultaneously preserving the knowledge learned from previously seen classes without the need to revisit old data. Specifically, we pretrain the detector using sufficient data from base datasets and then generate a set of class-wise prototypes that represent the intrinsic characteristics of the data. In the incremental learning session, we design a Hybrid Prototypical Contrastive (HPC) encoding module for learning discriminative representations. Furthermore, we develop a prototypical calibration strategy based on the Wasserstein distance to overcome the catastrophic forgetting problem. Comprehensive evaluations conducted with two aerial imagery datasets show that our InfRS effectively addresses the iFSOD issue in remote sensing imagery. Code will be released.
\end{abstract}


\section{Introduction}
\IEEEPARstart{O}{bject} detection stands as a crucial and prominent task in remote sensing image analysis. In recent years, deep convolutional neural networks have greatly advanced object detection tasks \cite{1-li2020object,2-li2022deep,3-cheng2016survey}. Despite these advancements, such approaches often rely heavily on large-scale training datasets, which are costly and time-consuming to acquire. To tackle this issue, the emerging approach of few-shot object detection (FSOD) promises to reduce the need for extensive labeled datasets \cite{4-antonelli2022few,5-kohler2023few,6-wang2020generalizing}. However, traditional FSOD methods typically require revisiting data from old (base) categories when expanding to new (novel) categories to achieve descent performance. This dependency introduces two main challenges: first, storing base training data from previous tasks raises concerns about computational costs and privacy issues; second, as earth observation technologies advance and interests expand, new categories continuously emerge. Traditional FSOD approaches lack the flexibility and scalability required to enroll new categories incrementally. In contrast, humans can quickly grasp new concepts from just a few examples without the need to revisit previously acquired knowledge. Thus, it is highly desirable for machine learning models to gain the capability to learn incrementally in practical applications. Driven by these motivations, we study the more challenging task of iFSOD in remote sensing imagery in this study.

Ideally, an iFSOD model should achieve two goals: (1) efficiently learn novel categories from a limited number of annotated samples in a continual manner; (2) maintain performance on previously existing categories without access to old data. This presents challenges on two fronts: first, data scarcity can lead to overfitting in novel categories; second, the catastrophic forgetting issue may result in a drastic performance drop in base classes when expanding to novel categories. Since data from previous tasks is no longer accessible when learning novel categories, the problem of forgetting is further exacerbated. Prior iFSOD works \cite{7-perez2020incremental,8-yin2022sylph,9-cheng2021meta} utilize a meta-learning strategy: the detector is initially pretrained on base categories, and meta-trained with few-shot episodes to gain the ability to enroll new categories in the subsequent meta-testing stage. Recently, inspired by the prevalent two-stage fine-tuning approach (TFA), several works \cite{10-feng2022incremental,11-zhang2023incremental} pretrain the base detector, and transfer the model on novel data. Nevertheless, existing iFSOD methods that are primarily developed for natural scene images tend to overlook the unique challenges presented by remote sensing imagery. Objects in remote sensing images typically have more complex backgrounds, diverse characteristics, arbitrary orientations compared to those in natural scenes \cite{12-ding2021object, 13-li2022deep}. These conditions cause insufficient discriminative feature learning, especially in a limited data scenario, rendering the models more vulnerable to the challenges of over-fitting and catastrophic forgetting. 

\begin{figure*}
	\centering
	\includegraphics[width=0.8\linewidth]{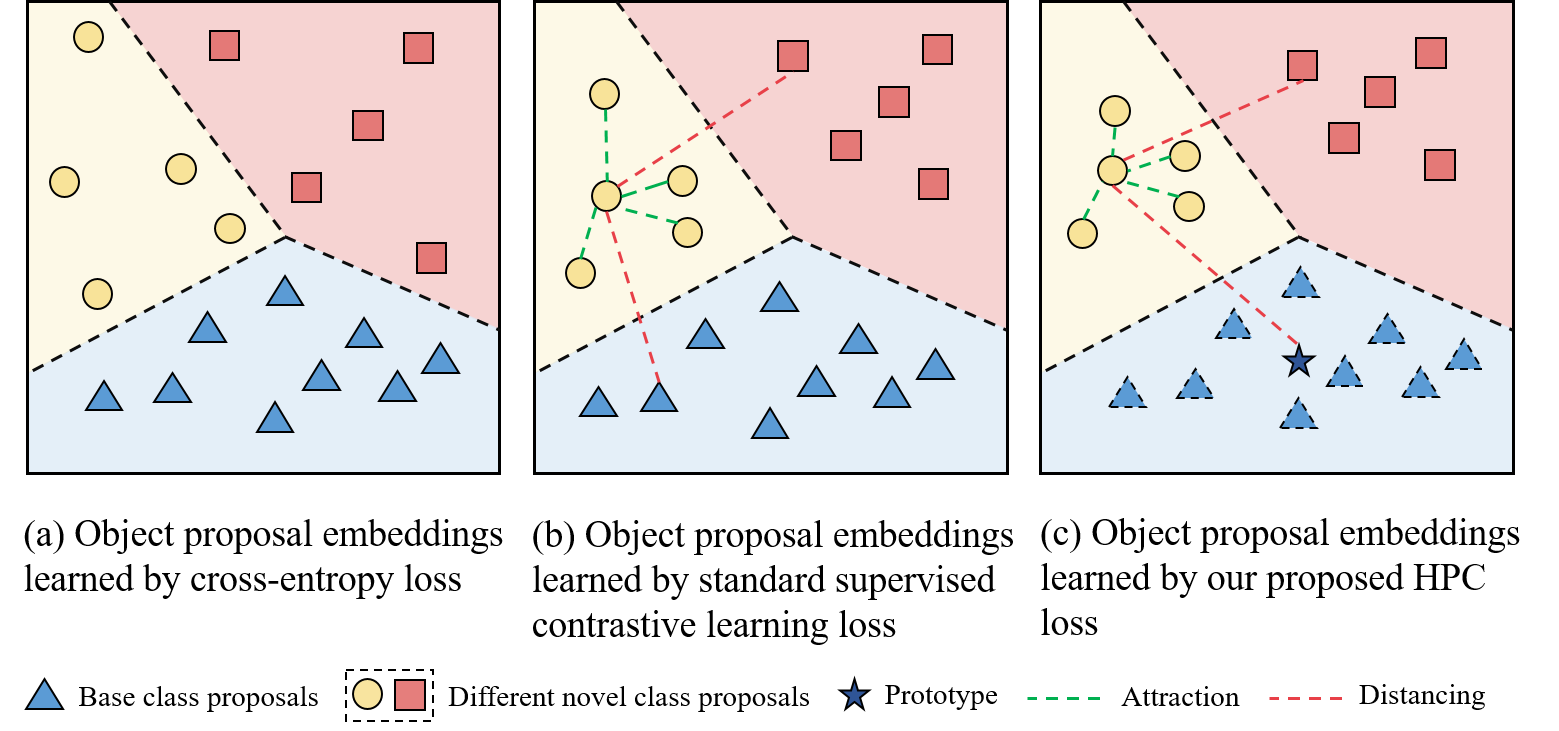}
	\caption{Comparisons between object proposal embeddings learned through different paradigms: (a) Object proposal embeddings learned by a general softmax classifier, which are not discriminative enough. (b) Standard supervised contrastive learning, which depends on instance-wise contrast to learn a discriminative feature distribution. (c) We propose an HPC encoding module that combines the advantages of both base class prototypes and novel instances. This provides explicit supervision to facilitate iFSOD.
 }\label{fig1}
\end{figure*}

To tackle these challenges, we argue that the key to iFSOD lies in sufficient discriminative feature learning. Contrastive learning is a powerful approach for learning discriminative features in various tasks~\cite{14-khosla2020supervised,15-zheng2021weakly}. The fundamental idea behind contrastive learning is to draw similar samples (positive pairs) closer in the latent space, while separating dissimilar ones (negative pairs). However, base class samples are not accessible in the incremental learning stage, which makes conventional contrastive learning impractical. To address this issue, we propose to learn prototypes for base categories as representatives. As illustrated in Figure~\ref{fig1}, rather than explicitly sampling positive and negative instances, we utilize class-wise prototypes as anchors for determining positive and negative pairs. Specifically, novel samples with the same labels are pulled together and pushed away from prototypes of base categories and all other novel categories. We introduce the HPC encoding module to leverage prototypical knowledge as reliable auxiliary supervision to foster more discriminative and robust representation learning for novel classes, while also preserving the intrinsic overall class distribution. %This module leverages prototypical knowledge to provide reliable auxiliary supervision in contrastive learning, without accessing data from base classes in the incremental learning stage. As illustrated in Figure~\ref{fig1}, rather than explicitly sampling positive and negative instances, we utilize class-wise prototypes as anchors for determining positive and negative pairs. Specifically, novel samples with the same labels are pulled together and pushed away from prototypes of base categories and all other novel categories. 
Moreover, we propose a prototypical calibration strategy, incorporating prototypes and Wasserstein loss to apply additional regularization on the model’s parameter update process, aiming to mitigate the catastrophic forgetting problem.

In summary, we propose InfRS, an incremental few-shot object detector in remote sensing imagery. To enhance representation learning for novel classes and to preserve existing knowledge, we introduce two novel components: the Hybrid Prototypical Contrastive (HPC) encoding module and the prototypical calibration strategy. We summarize our contributions as follows:
\begin{enumerate}
\item  Our approach enables learning of novel objects with limited examples incrementally and maintains detection performance on base classes without the need to revisit old data. To the best of our knowledge, this is the first study to tackle the iFSOD task in remote sensing images.
\item We introduce the HPC encoding module that captures the underlying structure and patterns within the data, thereby encouraging the model to learn discriminative and robust representations. 
\item We develop a prototype calibration strategy which leverages the prototypical knowledge and Wasserstein distance to eliminate the need for restoring or replaying old data without compromising performance on base categories.
\item Extensive experiments with the NWPU VHR-10 and DIOR benchmark datasets showcases the efficacy of our InfRS for iFSOD task in remote sensing images.
\end{enumerate}

\section{Related Work} 

\subsection{Object detection}
 Modern object detection models are typically grouped into two types: two-stage detectors \cite{16-girshick2014rich,17-girshick2015fast,18-ren2015faster,19-he2015spatial,20-dai2016r,21-lin2017feature} and one-stage detectors \cite{22-redmon2016you,23-liu2016ssd,24-lin2017focal,25-law2018cornernet,26-duan2019centernet}. Two-stage detectors initially extract candidate object proposals and then refine these proposals for precise classification and regression. %Prominent two-stage detectors include R-CNN \cite{16-girshick2014rich}, SPPNet \cite{19-he2015spatial}, Fast R-CNN \cite{17-girshick2015fast}, Faster R-CNN \cite{18-ren2015faster}, RFCN \cite{20-dai2016r}, and Feature Pyramid Network (FPN) \cite{21-lin2017feature}. 
 In comparison, one-stage detectors %such as YOLO \cite{22-redmon2016you}, SSD \cite{23-liu2016ssd}, RetinaNet \cite{24-lin2017focal}, CornerNet \cite{25-law2018cornernet}, and CenterNet \cite{26-duan2019centernet}, 
 bypass the proposal generation process and predict on feature maps in a single step. One-stage detectors generally have an edge in detection speed but may compromise in accuracy, especially for dense and small objects. Recently, transformer-based approaches like DETR \cite{27-carion2020end} and Deformable DETR \cite{28-zhu2020deformable} have shown promising performance, yet they still suffer from slow convergence issue. In this work, we opt for the classical two-stage detection architecture, Faster R-CNN, for iFSOD to secure a balanced compromise between performance and training efficiency. 

For object detection tasks in remote sensing images, researchers have also made significant efforts. Several studies, such as R2-CNN \cite{29-pang2019mathcal}, SSPNet \cite{30-hong2021sspnet}, SCRDet \cite{31-yang2019scrdet}, and SCRDet++ \cite{32-guo2018geospatial}, focus on detecting objects of varying scales. RICNN \cite{33-cheng2016learning} and \cite{34-li2017rotation} have developed rotation-invariant convolutional neural networks to handle objects with arbitrary orientations. More recent advancements, including RR-CNN \cite{35-liu2017rotated}, Polardet \cite{36-zhao2021polardet}, $S^2A$-Net \cite{37-han2021align}, RRPN \cite{38-nabati2019rrpn}, and \cite{39-fu2020rotation}, employ oriented bounding box representations to enhance detection of oriented objects in aerial images.

\subsection{Few-shot object detection}
%FSOD aims at detecting novel classes with limited annotated novel samples and abundant base classes samples. 
Prevailing FSOD models can be generally divided into two categories: meta-learning-based \cite{40-kang2019few,41-yan2019meta,42-xiao2022few,43-wang2019meta,44-zhang2021meta} and transfer-learning-based methods \cite{45-wang2020frustratingly,46-wu2020multi,47-sun2021fsce,48-fan2021generalized,49-qiao2021defrcn,50-cao2021few}.

Meta-learning-based methods obtain a meta-learner by constructing and solving a series of sub-tasks, then leverage the task-level knowledge for quick adaptation to novel classes. Kang et al. \cite{40-kang2019few} first propose FSRW, a meta-detector with a feature reweighting module that utilizes channel-wise attention to aggregate query and support features. Meta R-CNN \cite{41-yan2019meta} introduces an instance-level attention mechanism, focusing on each Region of Interest (ROI) features instead of the entire feature map. FSIW \cite{42-xiao2022few} builds on FSRW by developing a more complex feature aggregation module and training on a balanced dataset.
MetaDet \cite{43-wang2019meta} distinguishes itself by separating category-agnostic and category-specific parameter learning, which facilitating the detection of novel classes. Meta-DETR \cite{44-zhang2021meta} integrates meta-learning into the DETR framework to make predictions at the image level, moving away from region-based features. 

Transfer-learning-based methods start with the pioneering TFA \cite{45-wang2020frustratingly}, which pretrains a base detector and then fine-tunes only the last layers of classification and regression for new data. Notably, TFA outperforms more complex meta-detectors. Building on this framework, MPSR \cite{46-wu2020multi} enhances TFA by learning to detect multi-scale objects. FSCE \cite{47-sun2021fsce} introduces learning contrastive-aware representations to mitigate misclassification issues. Retentive R-CNN \cite{48-fan2021generalized} retains the RoI branch from the base detector, applying it as regularization during the fine-tuning stage to prevent performance degradation on base classes. Qiao et al. \cite{49-qiao2021defrcn} propose decoupling the training of classification and box regression to address the contradiction between these two tasks. FADI \cite{50-cao2021few} leverages the semantics of novel and associated base categories to enhance inter-class separability by disentangling the classifiers for novel and base categories. 

\subsection{Few-shot object detection in remote sensing images}
Although FSOD approaches have been extensively studied for natural scene images, fewer works have investigated remote sensing imagery. The pioneering work FSODM \cite{51-li2021few} introduces a meta-detector equipped with a Feature Pyramid Network (FPN) module to extract feature maps with different scales from support samples, along with a feature reweighting module to effectively learn meta-information from these maps. Another meta-detector P-CNN \cite{52-cheng2021prototype} learns prototype representations from support images and then fuses the information to improve region proposal generation. %Xiao et al. \cite{53-xiao2021few} design a self-adaptive attention network to leverage instance-level relations and minimize background inference. 
To mitigate the loss of knowledge from the pretraining stage, DH-FSDet \cite{54-wolf2021double} integrates an additional prediction branch for fine-tuning novel classes. Recently, Zhang et al. \cite{55-zhang2024few} propose ST-FSOD to solve the incompletely annotated instances issue existed in the few-shot learning process. However, most FSOD studies focus on detecting novel instance in a non-incremental manner. In this study, we seek to bridge this gap for remote sensing imagery.

\subsection{Incremental few-shot object detection}
Motivated by the advancements and limitations of FSOD, Pérez-Rúa et al. \cite{7-perez2020incremental} first explore the iFSOD task with the model ONCE. ONCE introduces a class-specific code generator through meta-learning to recognize novel classes. Similarly, Sylph \cite{8-yin2022sylph} employs a hyper-network, echoing ONCE's approach but decoupling localization from object classification. Feng et al. \cite{10-feng2022incremental} propose a novel method that attaches a new classification header for each emerging novel class, alongside a bi-path multi-class head to facilitate knowledge transfer. Rather than modifying the multi-class classifier, Zhang et al. \cite{11-zhang2023incremental} leverage responses of novel features and reorganizing base weights, aiming to transfer knowledge to novel classes that are relevant to base classes. Unlike prior studies, we propose a new architecture for iFSOD in remote sensing imagery, marking the first exploration of this task in this domain. %To account for the large intra-class variances of geospatial objects, we propose the HPC encoding module to exploit the semantic information within class prototypes, enhancing discriminative feature learning for improved performance on novel classes. Furthermore, we devise a prototypical calibration strategy to tackle the performance deterioration caused by the catastrophic forgetting problem. 

\subsection{Contrastive learning with prototypes}
Contrastive learning, also known as "learning by comparison," has proven effective in learning discriminative representations. Contrastive objectives are formulated to minimize distances between similar (positive) sample pairs and maximize distances between dissimilar (negative) pairs simultaneously in the embedding space. In self-supervised settings, positive pairs are obtained from differently augmented views of the same instance. In supervised settings, positive or negative pairs are determined with respect to an anchor instance depending on the class labels. For example, Li et al. \cite{56-li2020prototypical} introduce prototypical contrastive learning in an unsupervised setting, generating prototypes as data cluster representatives to encourage instances to project closer to their corresponding prototypes. Otsuki et al. \cite{58-otsuki2023prototypical} extend this concept to multi-modal language understanding tasks, learning data cluster prototypes from various domains to enable contrastive learning. Ouyang et al. \cite{57-ouyang2023pcldet} propose PCLDet to learn contrastive-aware features for fine-grained object detection.

% In this study, since the access to base data is denied in the incremental learning stage, the standard supervised contrastive learning cannot be straightforwardly implemented. To tackle this issue, we propose to learn prototypes as class representatives to serve as anchors when determining positive and negative pairs, and employ both prototypical knowledge and object samples in supervised contrastive learning to further enhance the learning of inter-class distinctions.

\subsection{Wasserstein distance}
It is often advantageous to establish a quantitative metric to measure the dissimilarity between two distributions. For example, Label Smoothing \cite{61-muller2019does}, Confidence Penalty \cite{62-pereyra2017regularizing}, and Knowledge Distillation \cite{63-hinton2015distilling} employ \(f\)-divergence to measure the difference in distributions. Despite its popularity, \(f\)-divergence focuses solely on the semantic relationships within the same class across two distributions and overlooks the structural information among difference classes. This is because it calculates each index of the output dimensions independently \cite{64-wang2022contrastive}. The Wasserstein distance \cite{66-frogner2015learning} is defined as the minimum cost required to convert one distribution into the other. Wang et al. \cite{74-wang2021normalized} introduce a Wasserstein distance-based metric to replace Intersection-over-Union (IoU) in detecting tiny objects. Rubner et al. \cite{65-rubner2000earth} adopt Wasserstein distance for image retrieval, as it allows for measuring the distance between partial matched distributions. 

In light of this, we employ the Wasserstein distance as a more suitable measure for computing distribution distances in iFSOD. As the incremental detector encounters an increasing number of novel classes, the Wasserstein distance can naturally be applied to compare its predicted distribution with its subset, which is predicted by the base detector and has partially matched indices.

\section{Methodology}
First, we introduce the problem setting for the iFSOD task. Subsequently, the framework for our InfRS is presented. Finally, the HPC encoding module and the prototypical calibration strategy are delineated.

\subsection{Problem definition}
In iFSOD, our goal is to develop a model capable of incrementally learning to detect novel categories using limited annotated examples, without revisiting or forgetting previously learned base classes. We utilize two disjoint datasets: $C_{\text{base}}$, which comprises abundant annotated data for base classes, and $C_{\text{novel}}$, which contains limited training data for each novel class. Initially, the model is pretrained on $C_{\text{{base}}}$ to establish a robust base detector. Subsequently, the base detector is expanded to include each novel classes in $C_{\text {novel}}$ incrementally through a fine-tuning process, without revisiting any data from $C_{\text {base}}$. The iFSOD detector is expected to be capable of effectively detecting all classes. Therefore, during the testing phase, we evaluate the overall performance using $C_{\text{all}} = C_{\text{base}} \cup C_{\text{novel}}$.

\subsection{Method overview}

\begin{figure*}
	\centering
	\includegraphics[width=\linewidth]{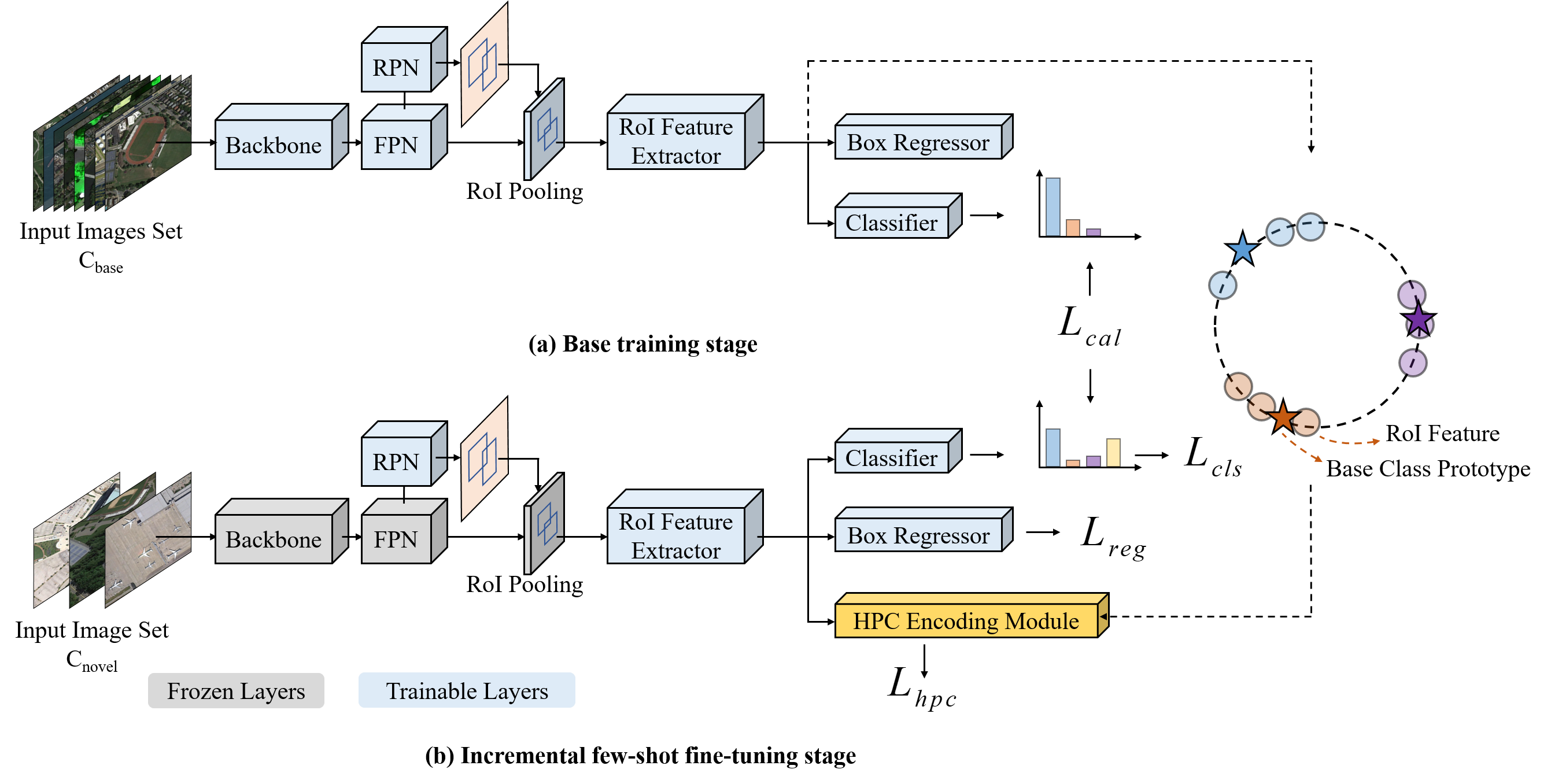}
	\caption{The overall framework of our InfRS. (a) Initially, the model is pretrained on base data, from which we extract prototypes as representatives for each base class. (b) During the fine-tuning stage, the few-shot novel data is incrementally enrolled. The HPC encoding module leverages prototypical knowledge to learn proposal embeddings with inter-class distinction and intra-class compactness. Furthermore, the prototypical calibration strategy, based on the Wasserstein distance, is introduced to ease the catastrophic forgetting effect.
 }\label{fig2}
\end{figure*}

Figure~\ref{fig2} depicts the overall architecture of our InfRS.  %In the Faster R-CNN detection pipeline, the backbone network processes input images to generates feature maps with potential region proposals that are predicted by the Regional Proposal Network (RPN). Proposals are then pooled into fixed-size feature maps and converted into RoI feature vectors by the RoI feature extractor, consisting of two fully connected (FC) layers. Finally, the classifier and regressor determine the categories and bounding box coordinates of any detected foreground objects. Figure~\ref{fig2} illustrates the overall architecture of our InfRS.
Inspired by the prevalent FSOD approach TFA, we employ a pretrain-finetuning scheme to achieve iFSOD in remote sensing images. We utilize the well-established Faster R-CNN with FPN as the base detector. In the pretraining stage, the base detector is trained on $C_{\text {base}}$; Subsequently, we compute the mean vector of RoI features for each class, using it as the prototype representation of that class. We also generate a corresponding pseudo classification probability distribution for each prototype. In the incremental learning stage, we freeze the parameters of class-agnostic backbone along with the FPN and fine-tune the model on $C_{\text {novel}}$. Since $C_{\text {base}}$ is inaccessible during this stage, we utilize the prototypical knowledge to provide the explicit supervision information, thus boosting the overall performance of iFSOD. %Specifically, we propose the HPC encoding module to leverage the base class prototypes in a contrastive manner to guide the model to build more discriminative object feature representations. Moreover, we utilize the base class prototypes to regularize the adaptation for novel classes, enabling the model to maintain the detection performance for pre-learned base classes.

\subsection{Prototype Generation}
We aim to learn robust prototypes containing high-level discriminative information for each base class. The prototype generation procedures are formulated as follows. We interpret the network function of base detector as a trainable feature learner \(f_{\theta}\) parameterized by \(\theta\), followed by the final classification and regression layers. Consider a dataset of \(N\) labeled examples: \(X = \{(x_1, y_1), \ldots, (x_N, y_N)\}\), where each \(x_i\) represents a \(D\)-dimensional feature vector from an instance in \(\mathbb{R}^D\), and \(y_i\) denotes its label, which ranges from 1to $K$. \(X_k\) is a subset which contains \(N_k\) instances with the class \(k\) in \(X\). The RoI feature of each instance is computed through \(f_{\theta} : \mathbb{R}^D \rightarrow \mathbb{R}^M\). Typically, \(M = 1024\) in Faster R-CNN framework. Based on these RoI features, we represent the class \(k\) with its prototype representation \({p}_{k} \):

\begin{equation} \label{equ1}
\begin{aligned}
    {p}_{k}=&\frac{1}{|N_k|} \sum_{(x_i,y_i)\in {X_k}}f_{\theta}(x_i).
\end{aligned}
\end{equation}

Assume we take the prototypes as the activation at the penultimate layer, and \({w}_{i} \) as the weight parameters of the softmax layer, then the input to the softmax layer \({o} \) can be written as
\begin{equation} \label{equ2}
\begin{aligned}
    {o}=\sum_{i=1}^{M}w_{i}p_{ki}.
\end{aligned}
\end{equation}
In the above equation, \({p}_{ki} \) represents the $i$-th component of the prototype \({p}_{k} \), and the predicted discrete probability distribution of prototype \({p}_{k} \) across $K$ classes can be formulated as:
\begin{equation} \label{equ3}
\begin{aligned}  
    {d}_{k}=\frac{exp(o_{k})}{\sum_{j=1}^{K}exp(o_{j})}.
\end{aligned}
\end{equation}

\subsection{Hybrid Prototypical Contrastive Encoding Module}
To integrate prototypical contrastive learning into the object detection framework, we design an FC layer of dimensions \(1 \times 128\) to encode the RoI features and class prototypes. Following this, we measure the similarity between the encoded vectors and optimize a contrastive objective to draw the vectors with the same labels closer while separating those with differing labels in the latent space. Specifically, for a novel instance, we learn a representation that increases the dissimilarity with the prototypes of base classes and other novel samples from different categories, while simultaneously minimizing the distance between samples belonging to the same class.  

We design our HPC loss function with considerations tailored for detection tasks. We denote \(\{z_{i},u_{i},y_{i}\}_{i=1}^{N}\) as the set of \(N\) RoI features in a mini-batch, where \(z_i\) represents the vector embedded by the HPC encoding module, \(u_i\) is the IoU score matching with its ground truth box, and \(y_i\) is the label. Let \(p = \{p_1, p_2, \ldots, p_K\}\) be a set of class prototypes projected by the HPC encoding module, with \(K\) denoting the total number of classes. We define the similarity score as $sim(\boldsymbol{z}_{i},\boldsymbol{z}_{j})=\frac{\boldsymbol{z}_{i}\cdot\boldsymbol{z}_{j}}{|\boldsymbol{z}_{i}||\boldsymbol{z}_{j}|}$, and the HPC loss function is formulated as follows:

\begin{equation} \label{equ4}
\begin{aligned}
\mathcal{L}_{HPC} = \frac{1}{N} \sum_{i=1}^{N} \mathbb I\{u_i\geqslant\phi\} \cdot L_{z_{i}},
\end{aligned}
\end{equation}

\begin{equation} \label{equ5}
\scriptsize
\begin{aligned}
&L_{z_{i}} =  \frac{-1}{N_{y_i}-1}\sum_{\substack{j=1,j\neq i}}^{N} \mathbb{I} (y_{i} = y_{j}) \\ 
& \cdot \log \left( \frac{\exp(sim(z_{i},z_{j}) / \tau)}{\left( \sum_{l=1, l \neq i}^{N} \exp(sim(z_{i},z_{l}) / \tau) + \sum_{k=1}^{K} \exp(sim(z_{i},p_{k}) / \tau) \right)} \right),
\end{aligned}
\end{equation}

where \(N_{y_i}\) is the total number of RoI features with the same label, \(\tau\) is a scalar temperature hyper-parameter as in \cite{14-khosla2020supervised} and $\mathbb{I} $ is an indicator function. $\mathbb I\{u_i\geqslant\phi\} $ represents an IoU threshold. We empirically set $\phi=0.7$ so that the most centrally RoI features are sampled to train the HPC encoding module. $\mathbb{I} (y_{i} = y_{j}) = 1$ if the $i$-th RoI feature and the $j$-th one belong to the same class, and 0 otherwise. Optimizing this loss function increases the distance between novel instances and base class prototypes, while simultaneously bringing instances of the same category closer in the embedding space. This results in tighter data clusters for each category, thus deriving well-separated decision boundaries and reducing misclassification on overall classes.

\subsection{Prototypical Calibration Strategy}
 The data scarcity setting in iFSOD not only leads to severe overfitting for novel classes but also results in catastrophic performance degradation on base classes. In this section, we present a prototypical calibration strategy to mitigate the catastrophic forgetting problem in the incremental learning phase. Specifically, we capitalize on the prototypical knowledge acquired during the base training stage to provide auxiliary prior information during the adaptation process for novel classes. We begin by extracting the prototype and preserving its output distribution as predicted by the base detector. We then encourage the prediction distribution of the fine-tuned model on the prototype approximating its original distribution. 

Concretely, consider two sets of distributions over a prototype denoted by \(p_{k}\). Let \(D=\{d_j | j=1,2,\ldots,K\}\) represent the set predicted by the base detector and \({F}=\{f_i | i=1,2,\ldots,Q\}\) represent the set predicted by the fine-tuned model, where \(f_i\) denotes the classification probability corresponding to class index \(i\), and similarly, \(d_j\) for class index \(j\). To calculate the Wasserstein distance between the two distributions, one must solve the transportation problem outlined below:

\begin{equation} \label{equ6}
\begin{aligned}
\operatorname*{minimize}_{T} \hspace{1cm} & \sum_{i=1}^{Q}\sum_{j=1}^{K}C_{ij}T_{ij} \\
\text{subject to}  \hspace{1cm} & T_{ij}\geq0, \quad i=1,\ldots,Q, \quad j=1,\ldots,K, \\
& \sum_{j=1}^{K}T_{ij}=f_{i}, \quad i=1,\ldots,Q, \\
& \sum_{i=1}^{Q}T_{ij}=d_{j}, \quad j=1,\ldots,K.
\end{aligned}
\end{equation}

In this equation, \(T\in\mathbb{R}^{m\times k}\) represents the transportation matrix and \(C\in\mathbb{R}^{m\times k}\) the cost matrix. $C_{ij}$ indicates the per-unit cost of transporting probability from category $i$ to category $j$. $T_{ij}$ represents the volume of probability transferred from category $i$ to category $j$. We employ the Sinkhorn algorithm \cite{67-ramdas2017wasserstein,69-altschuler2017near,70-cuturi2013sinkhorn} to compute the solution \( \tilde{T}\) to Equation~\eqref{equ6}, which can solve the optimal transportation problem with a linear convergence rate by imposing a straightforward structure on the transportation matrix. 

For $K$ prototypes representing the total $K$ base classes, we introduce a plug-and-play regularization term denoted as \(L_{cal}\), defined as:

 \begin{equation} \label{equ7}
 \begin{aligned}
 W_k = \sum_{i=1}^{Q}\sum_{j=1}^{K}\tilde{T}_{ij}\cdot C_{ij},
 \end{aligned}
 \end{equation}

 \begin{equation} \label{equ8}
 \begin{aligned}
 L_{cal}=\frac{1}{K}\sum_{k=1}^{K}W_k.
 \end{aligned}
 \end{equation}
 By optimizing \(L_{cal}\), we ensure the fine-tuned model predicts probability distributions similar to those of the base detector for the same prototype, thus preserving the knowledge acquired during the pretraining stage.

\subsection{Training Objectives}
The pretraining stage emloys Faster R-CNN loss function: ${L_{base}}=L_{cls}+L_{cls}+L_{rpn},$ where \(L_{cls}\) and \(L_{reg}\) are outputs of the last object classifier and regressor respectively, and \(L_{rpn}\) is a cross-entropy loss for generating object region proposals. When expanding to novel classes in the incremental learning stage, the contrastive loss \(L_{hpc}\) and the calibration loss \(L_{cal}\) are jointly optimized with ${L_{base}}$. To sum up, our loss function is defined as:
\begin{equation}\label{equ9}
L=L_{cls}+L_{reg}+L_{rpn}+\lambda_1L_{hpc}+\lambda_2L_{cal}.
\end{equation}
In the above formula, \(\lambda_1\) and \(\lambda_2\) are two factors designed to balance the importance of losses. 

\section{Experiments}
In this section, we begin by describing the datasets and outlining the implementation details. Subsequently, we present comprehensive results compared to the baseline model. We maintain consistency with iFSOD data settings and evaluation protocols to ensure a fair comparison. Finally, we conduct ablative studies to validate the effectiveness of the proposed components.

\subsection{Datasets}

1) \textbf{NWPU VHR-10}. This dataset comprises 10 geospatial object classes for detection: airplane, vehicle, ship, harbor, baseball diamond,basketball court, tennis court, ground track field, storage tank, and bridge. The spatial resolutions of the images in this dataset vary between 0.5 to 2 meters. % It contains a positive subset with 650 annotated images with objects from the specified 10 classes. 

2) \textbf{DIOR}. This dataset contains 23,463 aerial images with 192,472 object instances distributed into 20 categories: airplane, wind mill, , airport, train station, baseball field, tennis court, bridge, storage tank, chimney, ship, dam, expressway service area, stadium, expressway toll station,harbor, golf course, vehicle,ground track field, overpass, and basketball court. The spatial resolution varies between 0.5 and 30 meters, with each image measuring \(800 \times 800\) pixels in size. %The training, validation, and testing sets contain 5,862, 5,863, and 11,738 images, respectively.

To establish the data setup for iFSOD, we follow previous FSOD studies, except that the base data is only available in the pretraining stage and the novel categories are learned sequentially. In the NWPU VHR-10 dataset, the 10 classes are distributed among 3 novel categories and 7 base categories. We partition the positive dataset into two subsets: 70\% of the images (454 in total) are allocated for training, while the rest 30\% are designated for testing. Similarly, within the DIOR dataset, 15 categories are designated as novel, while the remaining 5 are considered base categories. %The model is trained on the training set and its performance is evaluated on the validation set.
We utilize 3 different splits to perform iFSOD, with the specific novel/base class split settings provided in Table~\ref{tab3} and ~\ref{tab5}. In the incremental learning stage, only $K$-shot instances from each novel category are randomly chosen for training We set $K = 1, 3, 5, 10$ for the NWPU VHR-10 dataset, and $K = 3, 5, 10, 20$ for the DIOR dataset to account for its larger scale. To quantitatively evaluate the performance, the mean Average Precision (mAP) at an IoU threshold of 0.5 is reported. We run inference on the overall datasets which include both base and novel categories, and report the performance of novel categories (nAP50), base categories (bAP50), and overall categories (mAP50) separately. 

% \begin{table}[htb]
% \caption{Three different base/novel classes split setting for NWPU VHR-10 dataset } \label{tab1}
% \resizebox{\columnwidth}{!}{%
% \begin{tabular}{ccccc}
% \hline
% Split & \multicolumn{3}{c}{Novel Set}                        & Base Set \\ \hline
% 1     & airplane         & baseball diamond   & tennis court & The rest \\
% 2     & ship             & storage tank       & vehicle      & The rest \\
% 3     & basketball court & ground track field & harbor       & The rest \\ \hline
% \end{tabular}%
% }
% \end{table}

% \begin{table*}[htb]
% \centering
% \caption{Three different base/novel classes split Setting for DIOR dataset }\label{tab2}
% \begin{tabular}{ccccclc}
% \hline
% Split & \multicolumn{5}{c}{Novel Set}   & Base Set \\ 
% \hline
% 1     & airport & bridge & ground track field & vehicle & storage tank & The rest \\
% 2     & airplane & baseball field & tennis court & train station & wind mill & The rest \\
% 3     & expressway service area & harbor & overpass & ship & stadium & The rest \\ \hline
% \end{tabular}
% \end{table*}

\subsection{Implementation Details}
We utilize the prevalent FSOD method TFA as the baseline model and adopt ResNet-101 with FPN as backbone network for all experiments. We use feature pyramids from $\mathrm{P}_2$ to $\mathrm{P}_6$. No data augmentation of any kind is applied in our analysis. The SGD optimizer is adopted, with a momentum of 0.9 and a weight decay set at 0.0001. Models are warmed up for the first 100 iterations. For the NWPU VHR-10 dataset, we pretrain the models with a batchsize of 4 for 24 epochs and fine-tune till full convergence. The initial learning rate is 0.001 and decreases tenfold at epoch 12 and 18 in the pretraining stage. During the incremental learning stage, the learning rate is 0.0001 and decreases tenfold at 4000 iterations. For the DIOR dataset, we set the batchsize at 8 and pretrain the models for 18 epochs and fine-tune till full convergence. The learning rate is 0.005 and decreases tenfold at epoch 11 and 14 in the pretraining stage. During the incremental learning stage, the learning rate is 0.0001 and decreases tenfold at 8000 iterations. Regarding hyperparameters, \(\tau\) is set to 0.1 in Equation~\eqref{equ5} , \(\lambda_1\) and \(\lambda_2\) are set to 0.5 in Equation~\eqref{equ9}. We train and test the models using an NVIDIA GeForce RTX 3090Ti GPU of 24 GB memory.

\subsection{Incremental Few-shot Detection Results}
\begin{table}[htb]
  \centering
  \setlength\tabcolsep{3pt}
    \caption{IFSOD performance on novel categories of the NWPU VHR-10 dataset.} \label{tab3}
    \resizebox{\columnwidth}{!}{%
    \begin{tabular}{@{}lcccccccc@{}}
      \toprule
      \multirow{1}{*}{\textbf{~~~Method} } & \multicolumn{4}{c} {Baseline TFA} & \multicolumn{4}{c}{InfRS}\\ 
      \midrule
     Novel Set 1  & 1-shot& 3-shot & 5-shot  & 10-shot  & 1-shot & 3-shot & 5-shot & 10-shot \\
        \midrule   
      airplane            & 0.11 & 0.41  & 0.51   & 0.62  & \textbf{0.23} & \textbf{0.41} & \textbf{0.51}    & \textbf{0.72}  \\
      baseball  diamond   & 0.18 & 0.42  & 0.67   & 0.85  & \textbf{0.19}   & \textbf{0.50} & \textbf{0.70}    & \textbf{0.87}   \\
      tennis court        & 0.21 & 0.38  & 0.44   & 0.60  & \textbf{0.45}   & \textbf{0.59} & \textbf{0.57}    & \textbf{0.63} \\
      \midrule
      mean                 & 0.17   & 0.40  & 0.54 & 0.64  & \textbf{0.32}  & \textbf{0.50}   &\textbf{0.60}   & \textbf{0.74} \\
      \midrule
       Novel Set 2   & 1-shot& 3-shot & 5-shot  & 10-shot  & 1-shot & 3-shot & 5-shot & 10-shot \\
      \midrule
      ship           & 0.09  & 0.43  & 0.54   & 0.66    & \textbf{0.09}    & \textbf{0.51}  & \textbf{0.64}  & \textbf{0.73}      \\
     storage tank    & 0.45  & \textbf{0.82}  & 0.72    & 0.74    & \textbf{0.51}    & 0.81  & \textbf{0.77}  & \textbf{0.82}      \\
      vehicle        & 0.09  & 0.14  & 0.15   & 0.13    & \textbf{0.09}    & \textbf{0.14}  & \textbf{0.18}  & \textbf{0.23}      \\
      \midrule
      mean           & 0.21   & 0.46   & 0.47  & 0.51  & \textbf{0.23}  & \textbf{0.49} & \textbf{0.53}     & \textbf{0.59}  \\
      \midrule
       Novel Set 3   & 1-shot& 3-shot & 5-shot  & 10-shot  & 1-shot & 3-shot & 5-shot & 10-shot \\
      \midrule
      basketball court    & 0.09   & 0.21   & 0.27  & 0.35   & \textbf{0.24}   & \textbf{0.28} & \textbf{0.37}     & \textbf{0.54}    \\
      ground track field  & \textbf{0.21}   & 0.46   & \textbf{0.44}  & 0.51   & 0.19   & \textbf{0.50} & 0.42     & \textbf{0.58}   \\
      harbor              & 0.17   & \textbf{0.03}   & \textbf{0.09}  & 0.27   & \textbf{0.17}   & 0.02  & 0.07    & \textbf{0.28}    \\
      \midrule
      mean                & 0.16   & 0.23   & 0.27  & 0.38   & \textbf{0.20}  & \textbf{0.27}  & \textbf{0.29}  & \textbf{0.47}  \\
      \bottomrule
    \end{tabular}%
    }
\end{table}

\begin{table}[htb]
  \centering
  \setlength\tabcolsep{3pt}
  \caption{IFSOD performance on base categories of the NWPU VHR-10 dataset of split 1.}  \label{tab4}
  \resizebox{\columnwidth}{!}{%
  \begin{tabular}{ @{}lccccccccccccccccc@{} }
    \toprule
    \multirow{2}{*}{\textbf{~~~Method} } & \multicolumn{4}{c}{{Baseline TFA}} & \multicolumn{4}{c}{InfRS}  \\
    \cline{2-5}                                     \cline{6-9}
    & 1-shot  & 3-shot   & 5-shot     & 10-shot       & 1-shot & 3-shot    & 5-shot     & 10-shot        \\
    \midrule
    ship                  & 0.90   & 0.91   & 0.91    & \textbf{0.91}  & \textbf{0.90}  & \textbf{0.91}  & \textbf{0.91}  & 0.90 \\
    storage tank          & 0.89   & 0.89   & \textbf{0.90}    & 0.89  & \textbf{0.95}  & \textbf{0.89}  & 0.89  & \textbf{0.90} \\
    basketball court      & \textbf{0.98}   & 0.61   & 0.43    & 0.61  & 0.97  & \textbf{0.97}  & \textbf{0.97}  & \textbf{0.95} \\
    ground track field    & 0.88   & \textbf{0.89}   & \textbf{0.91}    & 0.81  & \textbf{0.88}  & 0.86  & 0.86  & \textbf{0.89} \\
    harbor                &\textbf{0.88}   & \textbf{0.88}   & 0.84    & 0.86  & 0.82  & 0.84  & \textbf{0.88}  & \textbf{0.87} \\
    bridge                & 0.84   & \textbf{0.85}   & \textbf{0.88}    & 0.81  & \textbf{0.88}  & 0.78  & 0.85  & \textbf{0.82} \\
    vehicle               & 0.88   & \textbf{0.88}   & \textbf{0.89}    & 0.89  & \textbf{0.92}  & 0.87  & 0.87  & \textbf{0.89} \\
    \midrule
    mean                  & 0.89 & 0.84    & 0.82  & 0.82 & \textbf{0.90}  & \textbf{0.87} & \textbf{0.89}  & \textbf{0.89} \\
    \bottomrule
  \end{tabular}%
  }
\end{table}

\begin{table}[htb]
  \centering
  \setlength\tabcolsep{3pt}
    \caption{IFSOD performance on novel categories of the DIOR dataset.} \label{tab5}
    \resizebox{\columnwidth}{!}{%
    \begin{tabular}{@{}lcccccccc@{}}
      \toprule
      \multirow{1}{*}{\textbf{~~~Method} } & \multicolumn{4}{c} {Baseline TFA} & \multicolumn{4}{c}{InfRS}\\ 
      \midrule
     Novel Set 1  & 3-shot  & 5-shot   & 10-shot     & 20-shot       & 3-shot &5-shot    & 10-shot     & 20-shot \\
        \midrule   
      airport            & 0.07    & 0.15    & 0.13   & 0.18    & \textbf{0.10}    & \textbf{0.16}   & \textbf{0.15}   & \textbf{0.21}     \\
      bridge             & \textbf{0.09}    & 0.09    & 0.09   & 0.12    &0.08     & \textbf{0.15}   & \textbf{0.18}   & \textbf{0.18}     \\
      ground track field & 0.24    & \textbf{0.31}    & 0.27   & 0.35    & \textbf{0.28}   & 0.27   & \textbf{0.34}   & \textbf{0.37}     \\
      vehicle            & 0.16    & 0.15    & 0.29   & 0.35    & \textbf{0.22}    & \textbf{0.20}   & \textbf{0.35}   & \textbf{0.35}     \\
      storage tank       & 0.57    & 0.67    &0.67    & 0.67    & \textbf{0.66}    & \textbf{0.68}   & \textbf{0.67}   & \textbf{0.67}     \\
      \midrule
      mean               &0.23   & 0.27    & 0.29   &0.33     & \textbf{0.27}    & \textbf{0.29}   & \textbf{0.34}     & \textbf{0.35}     \\
      \midrule
       Novel Set 2   & 3-shot  & 5-shot   & 10-shot     & 20-shot       & 3-shot &5-shot    & 10-shot     & 20-shot \\
      \midrule
      airplane          &0.32     & 0.56   &0.58    & 0.67   &\textbf{0.48}    & \textbf{0.69}    &\textbf{0.74}     & \textbf{0.74}         \\
      baseball field    &0.60     &0.49    &0.68    &0.78    &\textbf{0.78}    &\textbf{0.71}    &\textbf{0.81}     & \textbf{0.81}         \\
      tennis court      &0.65     &0.60    &0.70    & \textbf{0.77}    & \textbf{0.75}   & \textbf{0.69}    & \textbf{0.76}    & 0.76        \\
      train station     &\textbf{0.03}      &0.11    & 0.16   &0.15    & 0.02  & \textbf{0.11}    & \textbf{0.16}    & \textbf{0.19}         \\
      wind mill         &0.15     & \textbf{0.18}    & 0.19   & 0.27   & \textbf{0.16}   & 0.16   &\textbf{ 0.19}    & \textbf{0.29}         \\
      \midrule
      mean             &0.35   & 0.39    & 0.46      & 0.53   & \textbf{0.44}  & \textbf{0.47}    &\textbf{0.53}     &\textbf{0.56}     \\
      \midrule
       Novel Set 3   & 3-shot  & 5-shot   & 10-shot     & 20-shot       & 3-shot &5-shot    & 10-shot     & 20-shot \\
      \midrule
      expressway service area  & 0.08   & \textbf{0.14}  & \textbf{0.18}    & \textbf{0.21}     & \textbf{0.13}   & 0.10   & 0.12   & 0.20     \\
      harbor                   & 0.03   & 0.07  & 0.07    & 0.12     & \textbf{0.03}   & \textbf{0.16}   & \textbf{0.12}   & \textbf{0.18}    \\
      overpass                 & 0.10   & \textbf{0.22}  & 0.25    & 0.30     & \textbf{0.17}   & 0.21   & \textbf{0.28}   & \textbf{0.30}      \\
      ship                     & \textbf{0.09}   & 0.09  & 0.12    & 0.15     & 0.04   & \textbf{0.26}   & \textbf{0.24}   & \textbf{0.27}     \\
      stadium                  & 0.29   & 0.24  & 0.48    & 0.40     & \textbf{0.29}   & \textbf{0.25}   & \textbf{0.49}   & \textbf{0.52}      \\
      \midrule
      mean                     & 0.13   & 0.15  & 0.22     & 0.23    & \textbf{0.13}    & \textbf{0.20}    & \textbf{0.25}     & \textbf{0.28}    \\
      \bottomrule
    \end{tabular}%
    }
\end{table}

\begin{table}[htb]
  \centering
  \setlength\tabcolsep{3pt}
  \caption{IFSOD performance on base categories of the DIOR dataset of split 1.}  \label{tab6}
  \resizebox{\columnwidth}{!}{%
  \begin{tabular}{ @{}lccccccccccccccccc@{} }
    \toprule
    \multirow{2}{*}{\textbf{~~~Method} } & \multicolumn{4}{c}{{Baseline TFA}} & \multicolumn{4}{c}{InfRS}  \\
    \cline{2-5}                                     \cline{6-9}
    & 3-shot  & 5-shot   & 10-shot     & 20-shot       & 3-shot &5-shot    & 10-shot     & 20-shot        \\
    \midrule
    airplane                      & 0.90    & \textbf{0.91}   & \textbf{0.91}   & 0.91    & \textbf{0.91}  & 0.90    & 0.90  & \textbf{0.91}  \\
    baseball field                & 0.79    & 0.90   & \textbf{0.89}   & \textbf{0.90}    & \textbf{0.90}  & \textbf{0.90}    & 0.81  & 0.89       \\
    basketball court              & \textbf{0.80}    & 0.73   & 0.79   & \textbf{0.80}    & 0.77  & \textbf{0.77}    & \textbf{0.79}  & 0.78       \\
    chimney                       & 0.09    & \textbf{0.82}   & 0.72   & \textbf{0.81}    & \textbf{0.82}  & 0.81    & \textbf{0.81}   & 0.77  \\
    dam                           & \textbf{0.33}    & 0.09   & 0.17   & 0.16    & 0.30  & \textbf{0.32}    & \textbf{0.31}  & \textbf{0.27} \\
    expressway   service area     & \textbf{0.59}    & 0.53   & 0.53   & 0.53    & 0.57  & \textbf{0.58}    & \textbf{0.59}  & \textbf{0.57}           \\
    expressway   toll station     & 0.43     & 0.53    & 0.53   & \textbf{0.54}    &\textbf{0.53}  & \textbf{0.53}    & \textbf{0.53} & 0.53           \\
    golf course                   & \textbf{0.57}     & \textbf{0.58}    & \textbf{0.56}    & 0.41    & 0.55  & 0.54    & 0.55  & \textbf{0.59}    \\
    harbor                        & \textbf{0.35}     & 0.22    & 0.22    & 0.16    & 0.32  & \textbf{0.30}    & \textbf{0.30}  & \textbf{0.30}   \\
    overpass                      & 0.22     & 0.07    & 0.09    & 0.06    & \textbf{0.27}  &\textbf{0.15}     & \textbf{0.22}  & \textbf{0.22}    \\
    ship                          & \textbf{0.80}     & 0.72    & 0.72    & 0.72    & \textbf{0.72}  & \textbf{0.72}    & \textbf{0.72}  & \textbf{0.72} \\
    stadium                       & 0.52     & 0.54    & 0.27    & 0.36    &\textbf{0.80}  & \textbf{0.76}    & \textbf{0.79}  & \textbf{0.78}    \\
    tennis court                  & 0.82     & 0.82    & 0.81    & 0.82    & \textbf{0.82}  & \textbf{0.82}    & \textbf{0.82}  & \textbf{0.82}      \\
    train station                 & \textbf{0.29}     & 0.20    & 0.27    & \textbf{0.29}    & 0.14  & \textbf{0.25}    & \textbf{0.28}  & 0.28     \\
    wind mill                      & 0.70     & \textbf{0.71}    & \textbf{0.71}    & \textbf{0.71}    & \textbf{0.70}  & 0.70    & 0.70  & 0.70   \\
    \midrule
    mean                          & 0.55      &0.56    & 0.55    & 0.54    & \textbf{0.61}   & \textbf{0.60}   & \textbf{0.61}   & \textbf{0.61}  \\
    \bottomrule
  \end{tabular}%
  }
\end{table}

\begin{figure*}
	\centering
	\includegraphics[width=\linewidth]{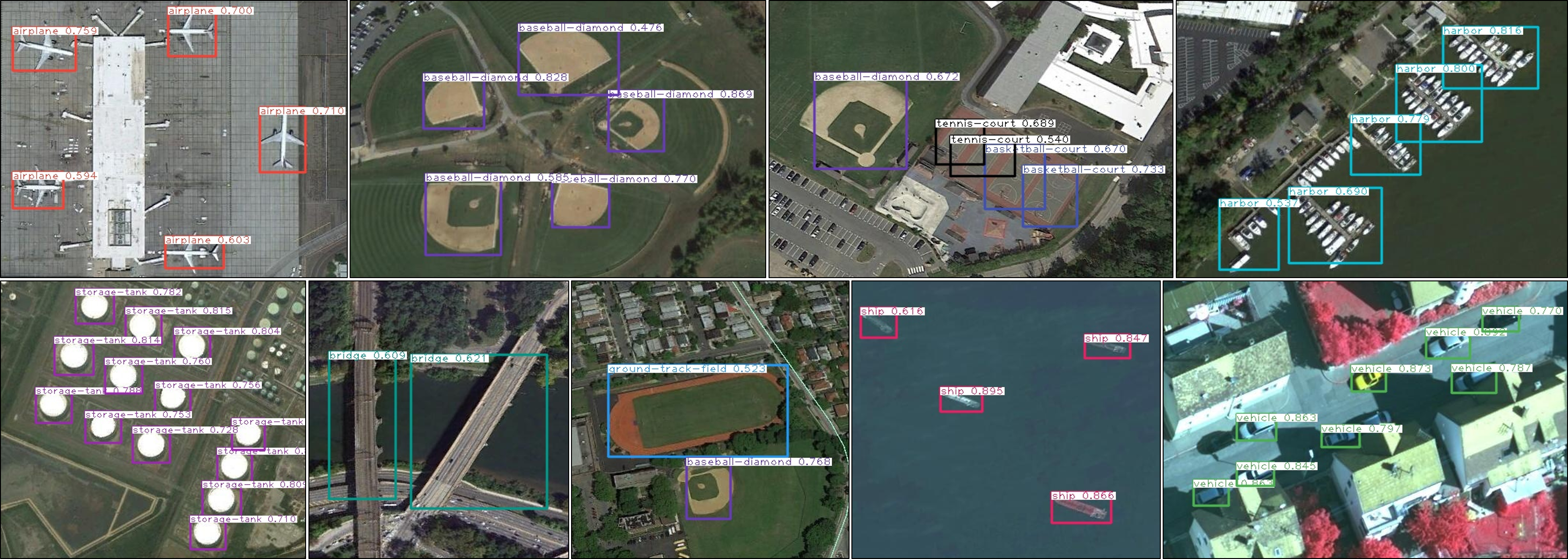}
	\caption{Visualization of the 10-shot detection results using our InfRS on the NWPU VHR-10 dataset for split 1. }\label{fig4}
\end{figure*}

\textbf{Results on NWPU VHR-10 dataset.} Table~\ref{tab3} and ~\ref{tab4} present the iFSOD results of both the baseline TFA and our InfRS on the NWPU VHR-10 dataset for both novel and base categories. As shown in these tables, our InfRS evidently outperforms the baseline method TFA in most scenarios. For example, considering the 10-shot setting, our approach obtains a mean mAP of $ 74\%$, $59\%$, and $47\% $ for novel classes across three different splits, which improve on the baseline method by $ 15.6\%$, $15.7\%$, and $23.7\%$, respectively. In the extreme low-data regime, such as 1-shot for tennis court, our InfRS achieves a remarkable leap in performance ($45\%$) compared to baseline TFA, demonstrating its robustness and generalizability. Regarding base classes, take split 1 for example, our model maintains the base class performance at $ 90\%$, $87\%$, $89\%$, and $89\% $ in $K = 1, 3, 5, 10$ shots, outperforming the baseline TFA by $ 1.1\%$, $3.6\%$, $8.5\%$, and $8.5\% $, respectively. Our model can effectively detect novel categories in incremental few-shot scenario and maintain the detection performance for base categories. We further present the qualitative results of our model in Figure~\ref{fig4}. 

\textbf{Results on DIOR dataset.}
The detection performance of the baseline TFA and our InfRS on the novel categories of the DIOR dataset is presented in Table~\ref{tab5}. Although the DIOR dataset is quite challenging due to objects of various scales and complex structures, our model effectively learn to detect novel categories without forgetting base categories. Taking split 1 as an example, our model achieves mean average precision (mAP) rates of 27\%, 29\%, 34\%, and 35\% with 3, 5, 10, and 20 examples, respectively, surpassing the baseline model by 17.4\%, 7.4\%, 17.2\%, and 6.1\%. For base classes, our model shows improvements of 10.9\%, 7.1\%, 10.9\%, and 13.0\% compared to the baseline under the same settings, as detailed in Table~\ref{tab6}. Equipped with the HPC encoding module and the utilization of prototypical calibration strategy, our method demonstrates its superiority in most iFSOD scenarios.

\begin{figure*}
	\centering
	\includegraphics[width=0.8\linewidth]{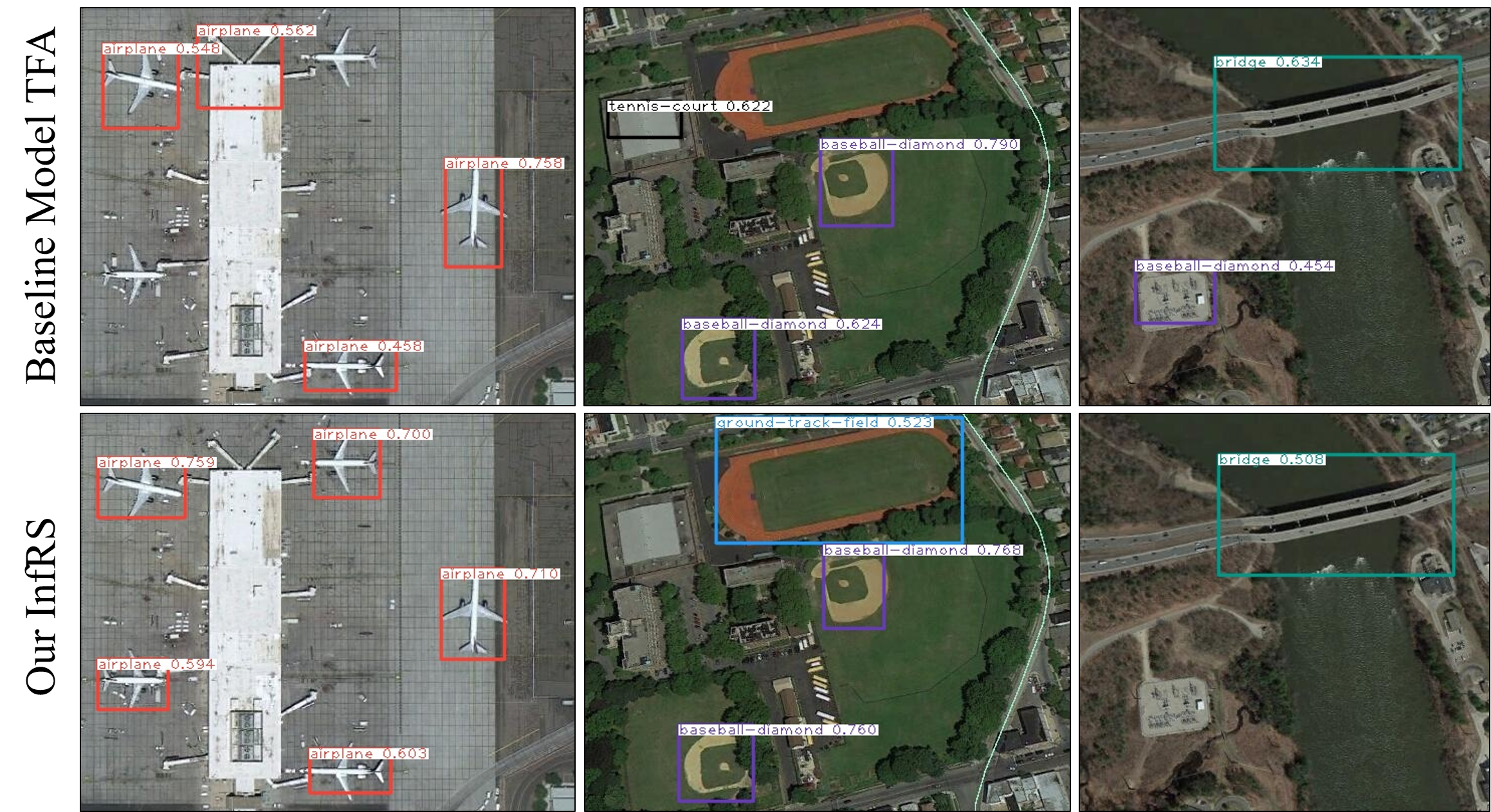}
	\caption{Qualitative comparison of the baseline model TFA and our InfRS. }\label{fig5}
\end{figure*}

\subsection{Ablation Study}
In this part, we conduct ablative experiments to carefully analyze the contribution of each component made to the ultimate performance of the InfRS model. All ablation results are reported for split 1 of NWPU VHR-10 dataset under a 10-shot scenario.

\begin{table}[htb]
\centering
\caption{Ablative experiments on the HPC encoding module and the prototypical calibration strategy proposed in our InfRS. \label{tab7}}
	\setlength\tabcolsep{1.5pt}%调列距
	\begin{tabular}{@{}cccc|ccc@{}}
 \toprule[1.5pt]
 % \Xhline{1.5pt}
		&\makecell[c]{fine-tune with \\ novel classes} & \makecell[c]{HPC loss} & \makecell[c]{Calibration loss} & bAP50   & nAP50   & mAP50   \\ 
 \midrule   
% \Xhline{0.75pt}

\textbf{a}  & \XSolidBrush & – & –  & 0.91 & 0.0 & 0.64 \\
\textbf{b}  & \CheckmarkBold        & – & –  & 0.82 & 0.64 & 0.77 \\
\textbf{c} & \CheckmarkBold & \CheckmarkBold &\XSolidBrush & 0.85 & 0.74 &0.82\\
\textbf{d} & \CheckmarkBold  &   \XSolidBrush   & \CheckmarkBold & 0.89 & 0.63 & 0.81\\
\textbf{e} (Ours) & \CheckmarkBold &  \CheckmarkBold & \CheckmarkBold  & 0.89 & 0.74 & 0.85\\
 \bottomrule[1.5pt]
 % \Xhline{1.5pt}
	\end{tabular}
\end{table}

% \begin{table}[htb]
% \centering
% \caption{Incremental few-shot detection performance under different combinations of $\lambda_{1} $ and $\lambda_{2} $}
% \label{tab8}

% \setlength\tabcolsep{4pt}%调列距
% 	% \begin{tabular}{@{}cc|ccc@{}}
% \begin{tabularx}{\linewidth}{>{\centering\arraybackslash}X>{\centering\arraybackslash}X|*{3}{>{\centering\arraybackslash}X}}
%  \toprule[1.5pt]
% \makecell[c]{$\lambda_{1} $}  &\makecell[c]{$\lambda_{2}$}  & nAP & bAP & mAP \\  
% \midrule   
% 0.1 & 0.1 &  0.70  &  0.90   &  0.84   \\
% 0.5 & 0.1 &  0.73  &  0.89   &  0.84   \\
% 0.9 & 0.1 &  0.74  &  0.89   &  0.85   \\
% 0.1 & 0.5 &  0.70  &  0.90   &  0.84   \\
% 0.5 & 0.5 &  0.73  &  0.89   &  0.84  \\
% 0.9 & 0.5 &  0.74   &  0.89   &  0.85   \\
% 0.1 & 0.9 &  0.70   &  0.90   &  0.84   \\
% 0.5 & 0.9 &  0.72   &  0.89   &  0.84   \\
% 0.9 & 0.9 &  0.72   &  0.89   & 0.84   \\ 
% \bottomrule[1.5pt]
% \end{tabularx}
% \end{table}

\textbf{Effectiveness of the proposed two components.} 
We investigate the effectiveness and contribution of the HPC encoding module and the prototypical calibration strategy proposed in our approach. We conduct ablative experiments in a systematic, five-step progressive manner:
\begin{itemize}
    \item \textbf{(a)} We pretrain the baseline model TFA using abundant data from base classes. Only the objective function of the Faster R-CNN framework is used.
    \item \textbf{(b)} We fine-tune the baseline model that has been trained on the base categories. The adaptation process employs the objective function used in the Faster R-CNN framework exclusively, without any additional loss terms.
    \item \textbf{(c)} We introduce the HPC encoding module and further fine-tune the model on novel categories. In the incremental learning stage, we utilize both the objective function utilized in the Faster R-CNN framework and the HPC loss term.
    \item \textbf{(d)} We employ the prototypical calibration strategy and fine-tune the base detector to learn novel categories. In this phase, we adopt the objective function utilized in the Faster R-CNN framework along with the calibration loss term.
    \item \textbf{(e)} We integrate both the HPC encoding module and the prototypical calibration strategy into the original model and fine-tune it for novel categories. In this stage, we jointly optimize both the HPC loss and the calibration loss proposed in our InfRS model, alongside the primary objective loss function of the Faster R-CNN framework.
\end{itemize} 

The results of these ablative studies affirm the efficacy of the two components proposed in our InfRS model. As reported in Table~\ref{tab7} \textbf{(a)}, the performance of the baseline TFA for novel classes is $0.0\%$, since novel data is not utilized in training. After adaptation for novel classes, the baseline model achieves an mAP of $64\%$, but the performance for base classes drops by $9.0\%$ due to the catastrophic forgetting effect. To counter the over-fitting problem, we introduce the HPC encoding module, which focuses on learning more discriminative features for novel classes. As the results of \textbf{(b)} and \textbf{(c)} in Table~\ref{tab7} shows, the performance reaches $74.0\%$ for novel classes, marking a $+10.0\%$ increase in nAP50 over the baseline model, validating the effectiveness of the HPC encoding module. Furthermore, this module contributes to a notable improvement ($+3.0\%$) for base classes, indicating that the enhanced classification accuracy for novel categories also benefits the learning of base categories. We conjecture that the base categories are also less prone to confusion with the novel categories, thereby further enhancing their accuracy. Analyzing the outcomes in \textbf{(b)} and \textbf{(d)}, the performance of base categories shows an enhancement of $+4.0\%$ in mAP50, highlighting the effectiveness of the prototypical calibration strategy in mitigating the catastrophic forgetting effect. Finally, equipped with the proposed two components, our InfRS achieves a satisfactory detection accuracy of $85.0\%$ for overall classes as indicated in Table~\ref{tab7} \textbf{(e)}. We further illustrate the comparative results of our InfRS against the baseline model TFA in Figure~\ref{fig5}.

% \textbf{Impact of loss weights $\lambda_{1} $ and $\lambda_{2}$.} We explore the influence of different combinations of loss weights $\lambda_{1} $ and $\lambda_{2}$ in Equation~\eqref{equ7} on incremental few-shot detection performance. Our experimental setup varies these weights among 0.1, 0.5, and 0.9. As shown in Table~\ref{tab8}, increasing $\lambda_{1}$ consistently improves performance for novel and overall classes, and the optimal balance achieves at $\lambda_{1} = 0.9 $ and $\lambda_{2} =0.1 $. 

\begin{figure*}
	\centering
	\includegraphics[width=\linewidth]{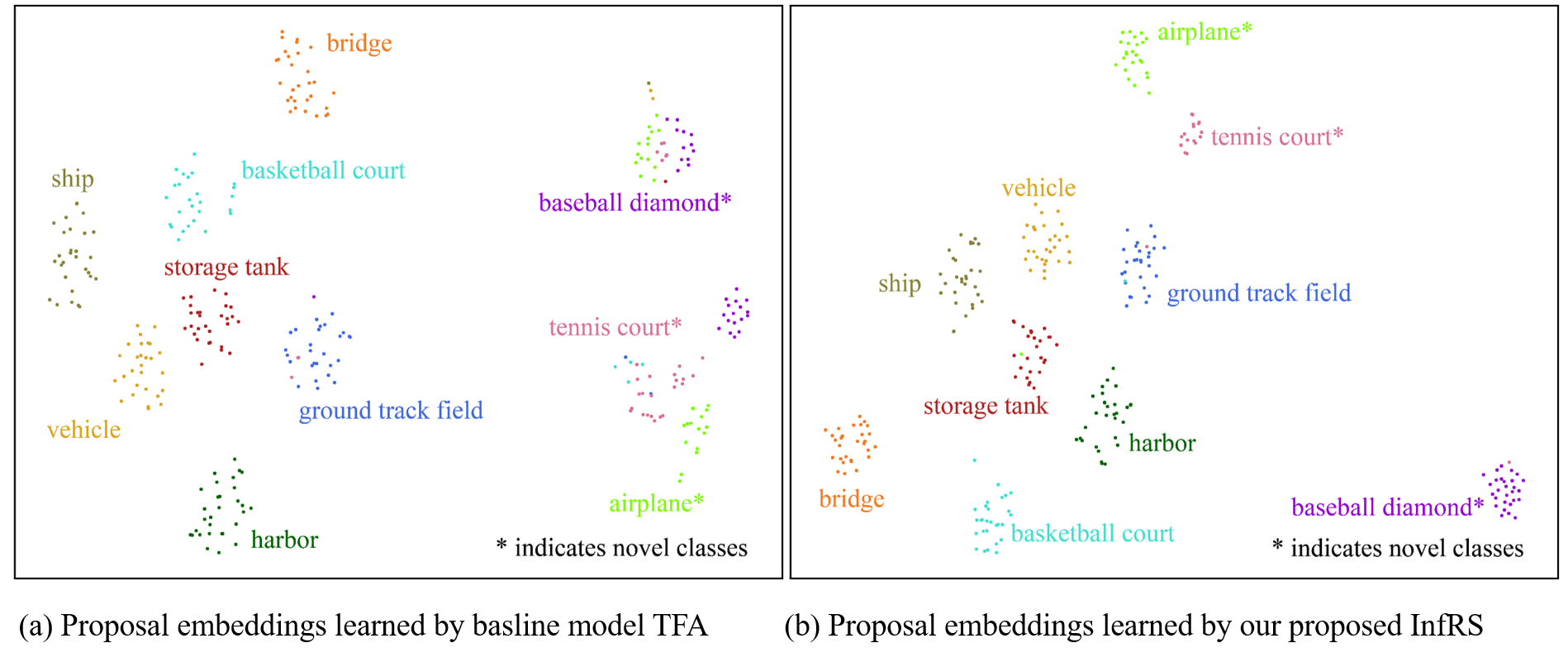}
	\caption{Visualization of the proposal embeddings learned by the baseline model TFA and our InfRS using t-SNE: (a) Overlap of the novel class clusters leads to sub-optimal decision boundaries and overlaps with the base classes, notably between the basketball court and tennis court categories. (b) In contrast, the decision boundaries between classes are well-defined, showcasing clear separation among both base and novel classes.}\label{fig3}
\end{figure*}

\textbf{Visualization and analysis.}  
To further validate the effectiveness of our InfRS, we visualize the embeddings of object proposals in the embedding space. The t-SNE \cite{71-Van2008visualizing} visualization of 30 samples per class learned from the NWPU VHR-10 dataset under a 10-shot scenario is presented in Figure~\ref{fig3}. We observe that our InfRS, powered by these two components, can learn well-defined boundaries between different classes.

\section{Conclusion}
This work investigates the challenging visual task of iFSOD in remote sensing imagery and presents a promising and novel solution named InfRS. We first formulate the iFSOD problem in remote sensing images and analyze the challenges associated with it. To address the overfitting issue, we propose the HPC encoding module to leverage both prototypical knowledge extracted from base categories and novel instances to learn contrastive-aware features. To address the catastrophic forgetting issue, we design the prototypical calibration strategy based on the Wasserstein distance to maintain the performance of base categories. Our InfRS effectively learns from data-scarce novel classes without revisiting or forgetting base classes, represents an initial foray into the domain of iFSOD in remote sensing.  

% \nolinenumbers

\bibliographystyle{IEEEtran}
\bibliography{ref}

% \vfill

\end{document}